\documentclass[letterpaper, 10 pt, conference]{ieeeconf}
\IEEEoverridecommandlockouts
\usepackage{graphics} 
\usepackage{epsfig} 
\usepackage{mathptmx} 
\usepackage{times} 
\usepackage{amsmath} 
\usepackage{amssymb}  
\usepackage{multirow}
\usepackage{booktabs}
\usepackage{float}
\usepackage{threeparttable}
\usepackage{bm}

\usepackage{cite}

\usepackage{color, soul}
\sethlcolor{blue}

\title{\LARGE \bf
Coarse-to-fine Semantic Localization with HD Map for Autonomous Driving in Structural Scenes
}

\author{Chengcheng Guo, Minjie Lin, Heyang Guo, Pengpeng Liang and Erkang Cheng$^{*}$ 
\thanks{Chengcheng Guo, Minjie Lin, Heyang Guo and Erkang Cheng are with the NullMax, Shanghai, 201210, China. Pengpeng Liang is with School of Information Engineering, Zhengzhou University, 450001, China.
        {\tt\small guochengcheng, linminjie, guoheyang, chengerkang@nullmax.ai; liangpcs@gmail.com}}%
\thanks{*Corresponding author.}
}

\begin{document}

\maketitle
\thispagestyle{empty}
\pagestyle{empty}

\begin{abstract}
Robust and accurate localization is an essential component for robotic navigation and autonomous driving. The use of cameras for localization with high definition map (HD Map) provides an affordable localization sensor set. Existing methods suffer from pose estimation failure due to error prone data association or initialization with accurate initial pose requirement. In this paper, we propose a cost-effective vehicle localization system with HD map for autonomous driving that uses cameras as primary sensors. To this end, we formulate vision-based localization as a data association problem that maps visual semantics to landmarks in HD map. Specifically, system initialization is finished in a coarse to fine manner by combining coarse GPS (Global Positioning System) measurement and fine pose searching. In tracking stage, vehicle pose is refined by implicitly aligning the semantic segmentation result between image and landmarks in HD maps with photometric consistency. Finally, vehicle pose is computed by pose graph optimization in a sliding window fashion. We evaluate our method on two datasets and demonstrate that the proposed approach yields promising localization results in different driving scenarios. Additionally, our approach is suitable for both monocular camera and multi-cameras that provides flexibility and improves robustness for the localization system.

\end{abstract}

\section{INTRODUCTION}

Vehicle localization (i.e., position and orientation estimation in the world coordinate system) is an important component of the autonomous driving system. For example, accurate and robust localization can provide useful information of decision making and perception module for autonomous driving.
Although the inertial navigation system based on IMU and RTK-GNSS can obtain enough accuracy, it is likely to fail in scenarios with poor GNSS signal, such as tunnels and skyscraper region \cite{jeong2019complex}. Also, it is not cost-effective for massive production. 
 
\begin{figure}[htb]
  \centering{\includegraphics[width=75mm]{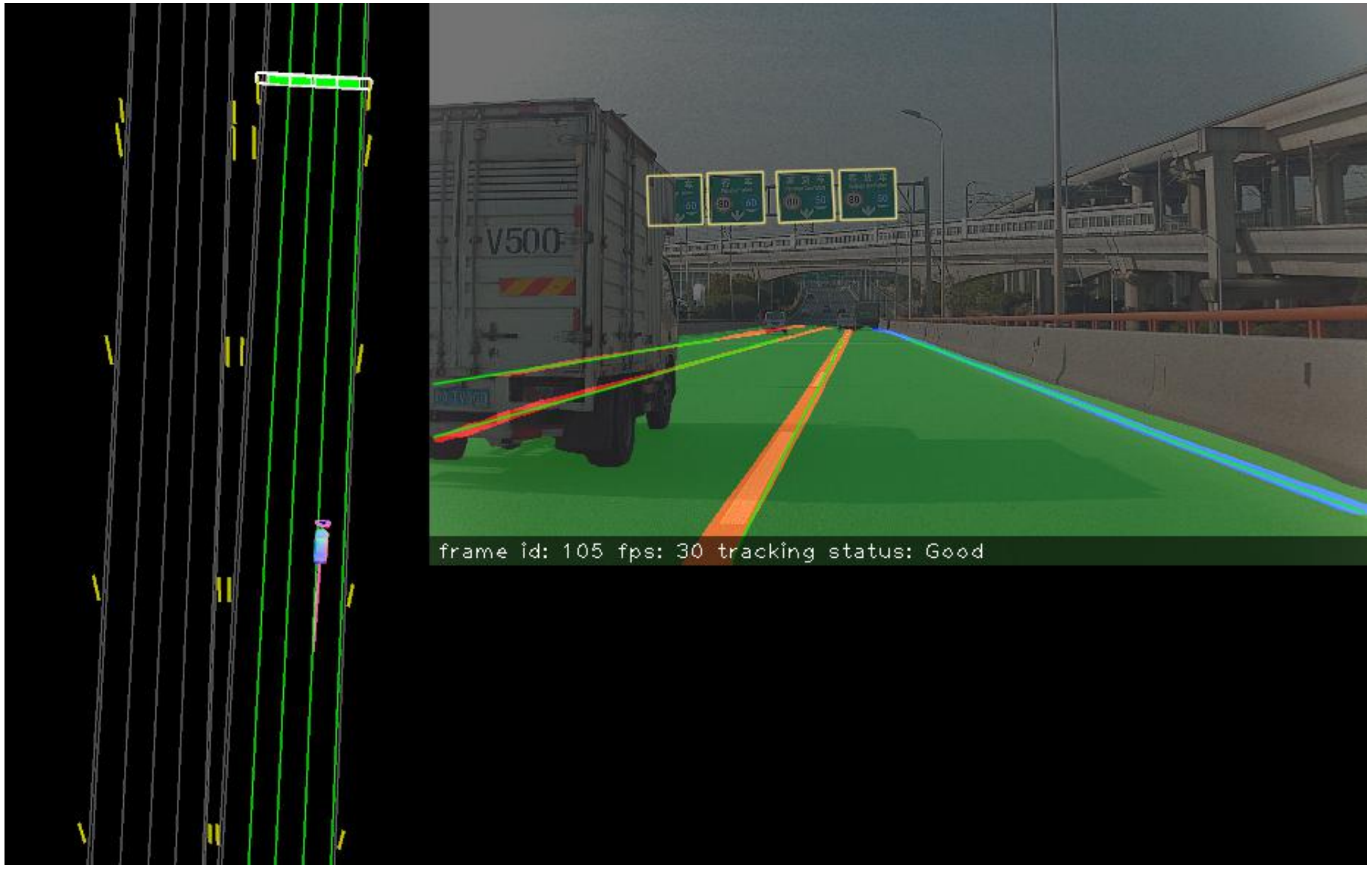}}\\
  \caption{Runing graph of vision localization with HD map. Video Link: https://youtu.be/rqWiGJevbcU} 
  \label{figure:localization in hdmap}
\end{figure}

In recent year, many works employ a offline-built prior map to solve the vehicle localization problem \cite{2020AVP} \cite{choi2020lane} \cite{paulsmonocular}. The prior map is usually applied as a semantic or geometric representation of the scene for autonomous driving. Lidar point \cite{zhang2014loam} and vision feature point \cite{mur2017orb} are two main categories of point cloud-based prior map. Although point-cloud map can provide sufficient localization accuracy, it is environmental sensitive and difficult to provide robust localization after a long temporal period. Also, point-cloud map can not scale due to its memory requirement. Compared to point-cloud map, vector-form HD map contains precious and rich semantic geometry information. HD maps are highly structured, organized as entities with geometry and attributes. Several works explore the localization method based on HDMap \cite{choi2020lane, paulsmonocular, xiao2020monocular}. However, data association is error prone or the localization system is not complete.

To address the aforementioned problems, the goal of our method is to provide a robust and accurate vision-based localization system. 
We introduce a coarse to fine vision localization by combining vector-form HD map and image semantic information. In system initialization step, a coarse initialization is provided by car-equipped GPS and then refined by exhaustive pose searching. 
In tracking stage, pose is estimated by aligning image semantic perception with landmarks of same semantic meaning in HD map. Specifically, given an image or multiple images, semantic segmentation result of entities in HD map is firstly obtained by deep learning method. Based on the segmentation result, a cost map is built by utilizing distance transform like function \cite{breu1995linear}. The minimization cost can be defined as projection photometric error of landmarks on the cost map. With additional wheel odometry information, the final pose is computed by pose graph optimization in sliding-window scheme. Finally, the lost recovery module is responsible for system re-initialization when failure happens in tracking stage. The proposed system running graph is shown in Figure. \ref{figure:localization in hdmap}.

To summarize, our main contributions are:

\begin{figure*}[htb]
  \centering{\includegraphics[width=170mm]{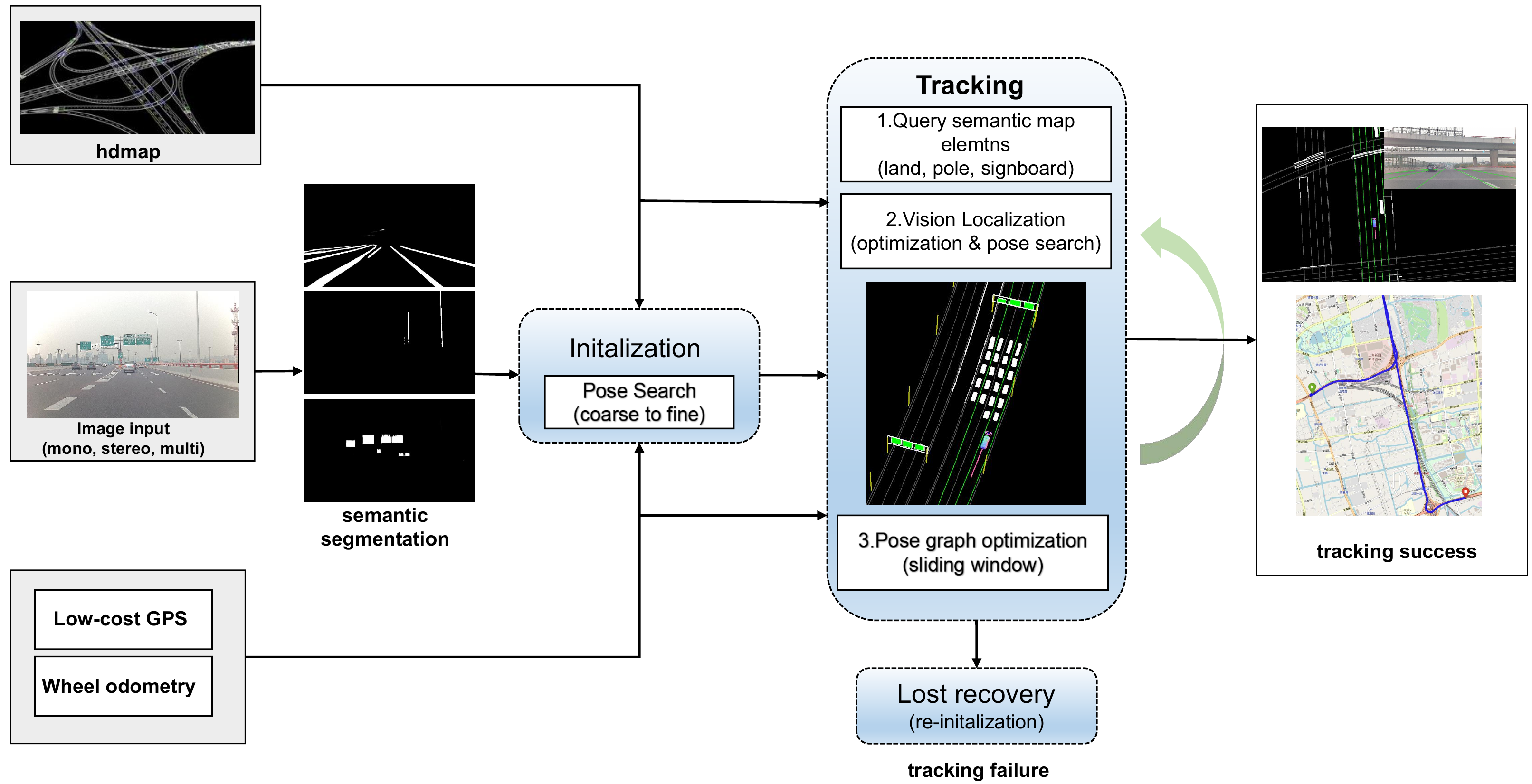}}\\
  \caption{Overview of the proposed vision localization system. Based on the prior map, low-cost GPS, wheel odometry input and camera signals, the 6-DOF pose can be estimated at centimeter-level accuracy.} \label{figure:pipeline}
\end{figure*}

\begin{itemize}

\item By leveraging semantic segmentation and HD map, we propose a complete vision localization system which includes initialization, tracking and lost recovery modules.

\item Our solution is flexible to handle both monocular camera and multi-camera system.

\item We evaluate our method on two datasets and demonstrate that our method yields promising localization results in different driving scenarios. 



\end{itemize}




\section{RELATED WORK}

Recently, there are many research works on vision-based localization with a prior environment map. 

\textbf{Point cloud map vs Vector-format map}
Prior map in localization can be categorized into pointcloud based map and vector-format map. The pointcloud map can be constructed by lidar or camera as sensors. Compare to point cloud map, compact vector-format HD Map is lightweight, easy to deploy and update. In order to extend the localization range, SuperPoint feature point map and semantic segmentation feature point map are both used for vehicle localization in \cite{livision}. By using lidar point cloud map, camera pose in monocular localization system, is calculated by finding correspondence between 2d lines in image and 3d lines in lidar map \cite{yu2020Monocular}. Hierarchical localization approach applies a different paradigm to finish vision localization task by introducing image retrieving into localization pipeline. For example, in \cite{sarlin2019coarse, sarlin2021back}, by matching the query image with the database images using global image feature, the correspondence between feature points of the image and the prior environment feature points map is obtained to estimate pose of query image. However, these methods are only suitable for indoor or small scale non-dynamic scenes. Many works exploits sparse semantic HD Maps in semantic localization \cite{paulsmonocular, ma2019exploiting}. Localization task is divided into two part in \cite{choi2020lane}: ego-lane identification and in-lane localization. The dash lane end point from map and image are used to localize vehicle to right position of the lane. However, only dashed lane used limits the localization application scenarios.

\textbf{Feature-based method} exploits low level geometry features or high level semantic features in the environment to build the association between image and map. Geometry feature includes point \cite{mur2017orb}, line \cite{yu2020Monocular} and plane \cite{yang2016pop}.  Camera pose can be estimated from the matching between extracted image feature and map. 
Examples of geometry features include ORB~\cite{rublee2011orb}, SIFT~\cite{ng2003sift}, etc. These geometry features provide distinguishing descriptions for matching task \cite{detone2018superpoint, zhang2013efficient}. However, they are not robust to environment variation, such as scenario changing from day to night or winter to summer. Therefore, they can not perform well for long-term localization. Compare to geometry features, semantics representation are alternative and widely used for localization in autonomous driving. Such representations include lane marking, curb, pole and so on \cite{schreiber2013laneloc, xiao2018monocular, xiao2020monocular, lu2017Monocular}.




\textbf{Direct method} does not require explicit keypoint detectors or feature descriptors. It can naturally sample pixels from across all image regions that have intensity gradient. For example, inter-frame pose is estimated based on the image alignment of image gradient points \cite{engel2017direct}. Recently, edge features are further used to produce distance transform image for pose optimization \cite{schenk2017robust, schenk2019reslam, kuse2016robust}.


Our approach is closely related to monocular localization with HD map \cite{paulsmonocular}. 
In \cite{paulsmonocular}, image features of elements in HD map are extracted by semantic segmentation. Distance transform operation is applied on binary segmentation result of each element in HD map to generate cost image for pose optimization. Finally, cost of re-projecting map elements on cost image according to initial pose is used to optimize the camera pose step by step. However, approach in \cite{paulsmonocular} is not a complete localization system which only concludes vision tracking module. The initialization step and lost recovering module, which are essential components for localization system, are not described. Its 6-DOF optimization strategy may produce estimation error when the image information is not enough to constrain vehicle pose. The proposed system supports multi-camera sensors setup, continuous localization is performed even when some of the cameras are blocked. Furthermore, semantic feature and robust strategy are used to make the system can run in the mapped environment with challenging conditions.

\section{Method}

Given an image or multiple images $I=\{I_i\}_{i=1}^N, N \geq 1$ captured from an autonomous driving system, with a HD map $M$, the vision-based vehicle global localization is to compute 6 DoF vehicle pose $\mathbf{T}_{wb} $. The map is defined as a set of meaningful landmark $M = \{ E_c \}_{c=1}^C$. The coordinate systems include: camera coordinate system $c$, vehicle baselink $b$ and HD map world coordinate system $w$, i.e. navigation coordinate system. Vehicle coordinate is FLU system, i.e. x axis points to forward direction, y axis points to left direction and z axis points to up direction.  Our framework consists of three major components: initialization, tracking and lost recovery. The system outputs 6 DoF vehicle pose relative to the map based on the input from camera, HDMap, low-cost GPS and wheel odometry. Fig. \ref{figure:pipeline} gives the overview of our framework.

\subsection{HD Map}

High precision map in autonomous driving, is usually a simple and flexible environment structure representative of the driving scenario. We use map elements $\{ E_i \}_{i=1}^M$ lane-markings (LA), pole-like objects (PO), signboards (SB) in vehicle localization. These elements are described by successive ordered three dimensional points collection in HD map. The graph in tracking part of Fig. 2 visualizes above mentioned semantic elements. In localization system, map elements can be queried by current vehicle position and a given search radius. For queried landmarks, we sampled points with a fixed length interval as landmark representative. 

\subsection{Semantic Segmentation and Post-Processing}

In order to find correspondence between HD map elements and image, semantic segmentation is applied to extract semantic features of image. We propose a lightweight deep learning network which can provide efficient segmentation results. Typically, the backbone is Resnet-18 \cite{he2016deep} and pre-trained on Cityscape dataset \cite{cordts2016cityscapes}. The network is a multi-head structure, each head is a binary segmentation of one element (LA, PO, or SB) in HD map for localization.

The vehicle pose estimation is achieved by non-linear optimization using semantic segmentation maps. We use different post-processing methods for semantic segmentation of different elements in HD map.
Given segmentation results of lane and pole, erode and dilate operations are used for gradient image generation.

\begin{figure}[htb]
  \centering{\includegraphics[width=85mm]{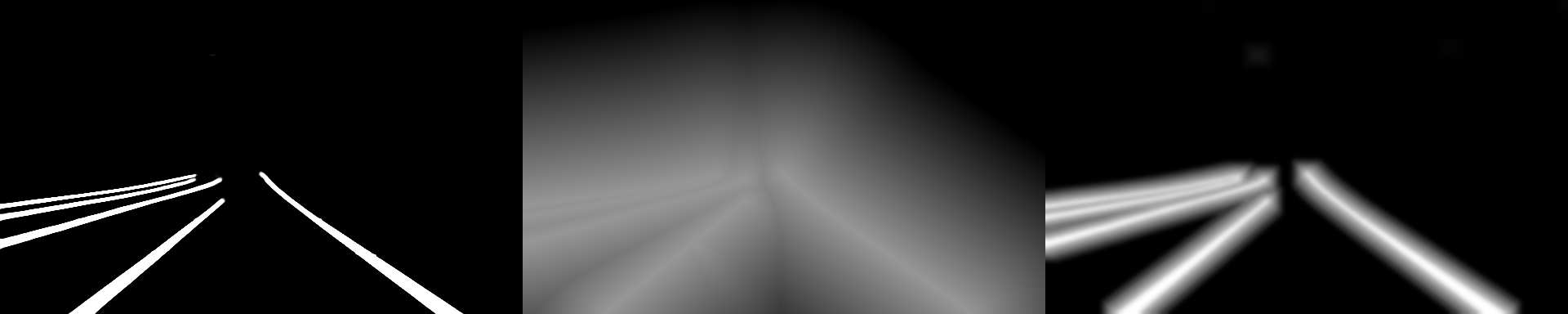}}\\
  \caption{Comparison between distance transform and morphology operation. Left: segmentation image of lane markings; Middle: cost map based on distance transform; Right: cost map with morphology operation.} \label{figure:dt_erode}
\end{figure}

For signboard landmark, Laplace transform is applied to extract edge information, then morphology operation is used to obtain smooth gradient image. Figure. \ref{figure:dt_erode} shows the difference of cost image between distance transform and morphology operation used by proposed method. The cost map generated by morphology is easier to make the pose optimization converge to the right result. Finally, the processed segmentation results are converted in the range of $[0, 1]$. We define the post-processed segmentation results by $I_s$.

\subsection{Initialization}
The purpose of initialization module is to obtain relative accurate pose estimation in map coordinate system for successive pose tracking step. We introduce a robust and accurate initialization method in a coarse to fine manner. Specifically, a coarse initial pose $ \mathbf{T}_{wb}$ is computed by two valid GPS records. Since vehicle can be in still status, the distance of two GPS point is set to a moderate value.
The $x$ and $y$ plane coordinate of vehicle are set to the second valid point. And $z$ coordinate is obtained based searched near map ground elements. Also, the roll angle $\theta_{x}$ and pitch angle $\theta_{y}$ of vehicle are set as zero. The yaw angle $\theta_{z}$ is set to the direction of two selected measurements. In order to get a high rate of successful initialization and more accurate initial pose result, the coarse initial pose is refined by exhaustive pose searching in a pre-defined grids.




The searching and optimization cost is defined by the sum of photometric residual \cite{engel2017direct} of all semantic landmarks, which can be written as: 


\begin{equation}
    cost = \sum_{i = 0}^n \|I_s(\pi((\mathbf{T}_{wb} * \mathbf{T}_{bc})^{-1}\mathbf{P}_w)) - 1.0 \|_2
\label{eqn:photometric error}    
\end{equation}

In Eqn. \ref{eqn:photometric error}, $\mathbf{P}_{w}$ is the 3D world coordinate of elements $\{E_i\}$ in the map $M$.
$\mathbf{T}_{bc}$ is the camera extrinsic parameters relative to vehicle baselink.
$\pi$ is projection function based on the camera model. 
We use different search parameters, search step and range, for different pose freedom. For example, the search step and range for vehicle lateral position are set to 0.2m and [-10m, 10m], which covers the error tolerance of car equipped GPS. Finally, the pose combination with minimum cost will be considered as the pose of the current initialization frame. The efficient implementation is achieved by CUDA acceleration.


\subsection{Tracking}

Given an initial pose, in tracking stage, vehicle pose is estimated based on the alignment between semantic feature and prior map. The tracking module can be divided into three steps. Firstly, vehicle pose $\mathbf{T}_{wb}^{k+1}$ of frame $k+1$ is predicted based on pose estimation $\mathbf{T}_{wb}^{k}$ at time $k$ and other sensor input such as vehicle wheel odometry measurement $ \mathbf{T}_{b}^{k \rightarrow k+1}$ by:

\begin{equation}
    \mathbf{T}_{wb}^{k+1} = \mathbf{T}_{wb}^{k} * \mathbf{T}_{b}^{k \rightarrow k+1}
\end{equation}


If the driving scenario meets the longitudinal constrain setting, a cropping local map from the global map step is performed. Otherwise, a longitudinal position correction process is applied first.

\textbf{Cropping local map from the global map} Map elements (LA, PO, and SB) are queried from the global map in a pre-defined short range using current coarse vehicle pose. Then the queried local map is applied for drift-free vision localization. Map element $E$ are projected back to image points $P$. In order to obtain an accurate pose optimization, points in $P$ are uniformly sampled in the image space.

\textbf{Longitudinal position correction} The longitudinal localization could suffer significant drift after a long period of time, in the case that the driving scenario does not meet the longitudinal constrains. For example, it happens when the queried lanes are (1) parallel to each other, (2) straight forward, and (3) there are no signboards or poles to restrict the vehicle's longitudinal translation in the environment. GPS signal is then used to update longitudinal localization of the vehicle pose  $ \mathbf{T}_{wb}^{k+1}$. Such longitudinal position correction mechanism is able to avoid the drift of the longitudinal localization in poor environmental conditions, especially for a long period of time.

Secondly, the 6 DoF vehicle pose is refined by image alignment with HD map elements. A cost map is built based image semantic segmentation and morphology operations. The alignment is solved by a non-linear optimization (Levenberg-Marquardt (LM) \cite{more1978levenberg}). In the case that there are missing vertical landmarks (e.g., signboards or poles) in the scene, $\theta_{y}$, $\theta_{z}$, $t_y$ and $t_z$ are computed by, $\theta_{y}$, $\theta_{z}$ and $t_y$ are firstly estimated and then $\theta_{y}$ and $t_z$ are optimized later. $\theta_{x}$ and $t_x$ are not included due to that the roll angle is usually very small when the vehicle is moving on a flat ground and longitudinal displacement of the vehicle is not observable when the vehicle and queried lanes are parallel to each other. In addition, to compromise that roll angle is missing in the optimization, vehicle rotation is then fine-tuned by a brute force search using a substantial range. Search interval of the rotation is set to 0.5 degree.

Lastly, in order to get a smoother pose for the planning module and to advance the robustness of the localization system, a pose graph is applied with a sliding window.
A well-tracked frame is included in the optimization window. If the window size exceeds a threshold, a frame from history will be excluded from the window  according to the vehicle state. For example, if the vehicle odometry measurement is close to zero, the second newest frame is picked, otherwise the oldest frame is used.

\begin{figure}[htb]
  \centering{\includegraphics[width=70mm]{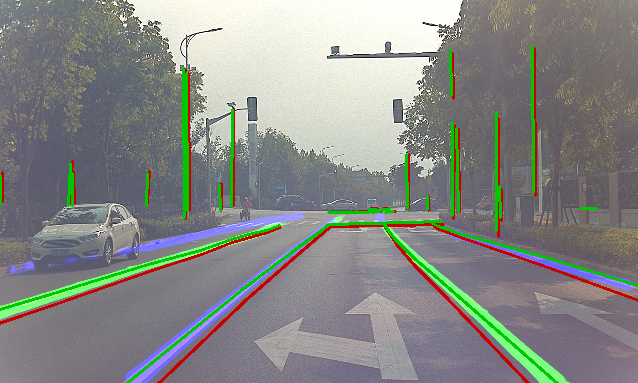}}\\
  \caption{Example of optimization by projecting HD map elements to an image. Initial pose projection is in red and optimization result is in green.} 
  \label{figure:before after}
\end{figure}

In pose optimization, the factor graph has two components. The first is the prior pose factor of each frame which constrains its prior distribution of vision alignment. The other is the wheel odometry factor which establishes the connection between the adjacent frames to ensure the smooth pose output. The total residual of pose graph optimization is shown in Eqn. \ref{eqn:total cost}. The G2O \cite{grisetti2011g2o} framework is used for the optimization process.

\begin{equation}
    \mathbf{T} = argmin \sum_{i, j \in w} \| ln(\mathbf{T}_{i}(\mathbf{T}_{i}^{*})^{-1})^{\vee} \|_{2} + \lambda \| ln(\mathbf{T}_{j}^{-1}\mathbf{T}_{i}\mathbf{T}_{ij})^{\vee} \|_{2}
\label{eqn:total cost}    
\end{equation}


$\mathbf{T}_{i}^{*}$ is the prior pose from semantic alignment and $\mathbf{T}_{ij}$ can be derived from measurement of wheel odometry. A localization confidence is computed in the pose estimation to evaluate the localization status. A lost recovery module is activated when the localization fails.

\subsection{Optimization}

Details about gradient of loss function are derived in following equations. Jacobian of   error relative to the optimized state is usually used to accelerate the process for non-linear optimization method (e.g., Gauss-Newton or LM):


\begin{equation}
\frac{\delta \text{error}}{\delta \epsilon} = \frac{\delta I_{s}}{\delta u} \frac{\delta u}{\delta p_c} \frac{\delta p_c}{\delta \epsilon},
\label{eqn:gradient}
\end{equation}
where \text{error} is the projection error of the cost map, ${\delta \epsilon}$ denotes the perturbation of vehicle pose, ${u}$ is the image coordinate and ${p_c}$ is the point in camera coordinate system. In order to support multi-camera observations, that optimization state is vehicle pose rather than camera pose.
The camera extrinsic parameter $\mathbf{T}_{bc}$ is applied for the transform between vehicle and camera coordinate systems. Camera extrinsic parameters are not included in optimization state. The last term of Eqn. \ref{eqn:gradient} when estimated state is 6 DoF pose can be further written as:

\begin{equation}
\frac{\delta p_{c}}{ \delta\epsilon} = \frac {\delta \mathbf{T}_{cb}(\mathbf{T}_{wb} * Exp(\delta\epsilon))^{-1}\mathbf{P}_{w}}{\delta \epsilon}
\label{eqn:cheap}
\end{equation}


\begin{equation}
\frac{\delta p_{c}}{\delta \epsilon} = -\left [ I_{3} \ -[p_{c}]_{\times} \right] Ad(\mathbf{T}_{cb}),
\label{eqn:cheap}
\end{equation}
where $[p_{c}]_{\times}$ is skew-symmetric matrix of $p_{c}$ and $Ad(\mathbf{T}_{cb})$ is the adjoint matrix of $\mathbf{T}_{cb}$. 

\begin{figure}[htb]
  \centering{\includegraphics[width=80mm]{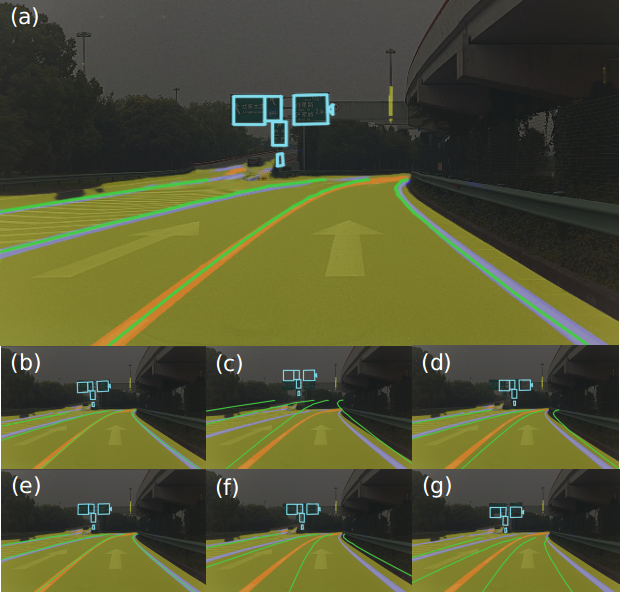}}\\
  \caption{
  Examples of alignment between HD map and image semantic segmentation. (a) An accurate alignment; (b)-(g) Alignment results under perturbation of freedom: roll, pitch, yaw, x, y and z. Semantic segmentation: freespace (yellow), lane markings (organ and purple). Projection from HD map: lane markings (green) and signboard (cyan).} \label{figure:pose vibration}
\end{figure}

In order to support optimization with multi-camera observations, vehicle pose rather than the camera pose is used as the optimization state. Also, Lie Algebra is used in the transform representation for the 6-DOF pose optimization. Nonholonomic pose freedom estimation including
3 DoF pose estimation and 2 DoF pose estimation exist in the localization system. Therefore the jacobian of photometric error relative to the vehicle pose expressed by Euler angle and translation needs to be derived.

\begin{equation}
    p_{c} = \mathbf{R}_{cb}((\mathbf{R}_{wb})^{-1}(\mathbf{P}_{w} - \mathbf{t}_{wb})) + \mathbf{t}_{cb}
\end{equation}

$\mathbf{R}_{wb}$ and $\mathbf{t}_{wb}$ are the rotation part and translation part of vehicle pose respectively. The translation part of jacobian is list in Equation. \ref{eqn:jt}.
\begin{equation}
    \frac{\delta p_{c}}{\delta t_{wb}} = - \mathbf{R}_{cb}(\mathbf{R}_{wb})^{-1} = -\mathbf{R}_{cw}
\label{eqn:jt}
\end{equation}

$\mathbf{R}_{wb}$ is represented in Z-Y-X rotation order. The jacobian of $p_{c}$ relative to yaw can be derived. 



\begin{equation}
    \frac{\delta p_{c}}{\delta \theta_z} = \mathbf{R}_{cb} \frac{\delta (\mathbf{R}_{wb})^{T}}{\delta \theta_z}(\mathbf{P}_{w} - \mathbf{t}_{wb})
\end{equation}
where the $\theta_{x}$ and $\theta_{y}$ are the same principle as $\theta_{z}$. Based on above derived jacobian, separate Euler angle and translation part can be optimized. All optimization process are divided into two steps. The first step is the optimization with robust kernel in order to suppress outliers. The second step is the optimization without robust kernel to obtain higher estimation accuracy by removing observations with large error in first step. Optimize result of single image is shown in Figure. \ref{figure:before after}.

\subsection{Lost Recovery}
However, the system may be lost in the following three situations: (1) vehicle is out of the operation domain of the HD map; (2) The total number of pose optimization failure exceeds a threshold; (3) The number of consecutive frames with severe occlusion exceeds a threshold (e.g, This happens in situation of traffic jam in which semantic map elements are totally invisible). The tracking confidence calculation module will determine system status based on above statistical indicators. When localization system is in lost status, lost recovery mode is activated. The pose of a lost frame is replaced by a back-up pose which is inferred from the wheel odometry, i.e. the pose before optimization. Given the next frame, in order to activate the tracking stage, the system turns into the initialization status again.

\vspace{5mm}
\begin{figure*}[htb]
  \centering{\includegraphics[width=170mm]{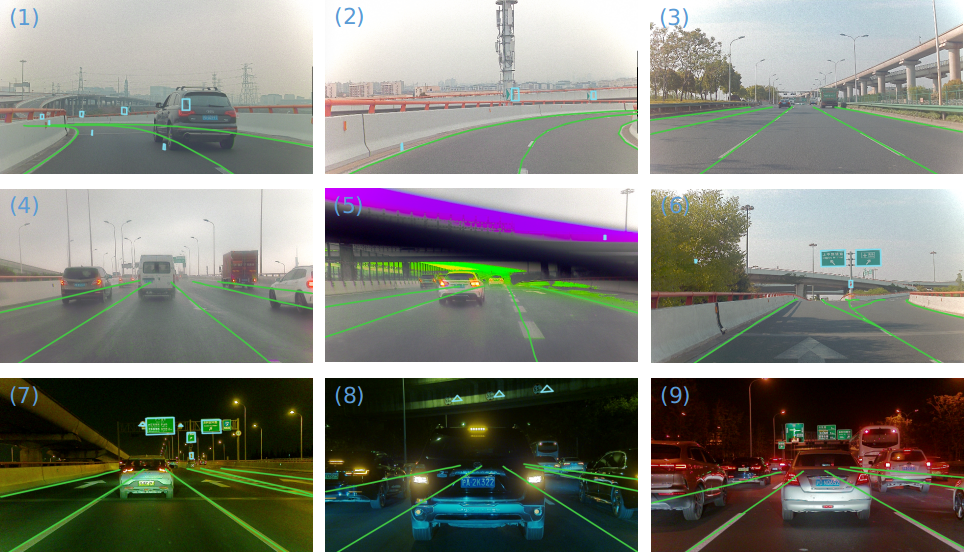}}\\
  \caption{Qualitative results on Shanghai dataset. Projection results of different scenes: (1-2) curve road; (3) long straight road in sunny day; (4) rainy; (5) windshield wiper blocks part of the image; (6) diverging ramp; (7) low illumination; (8-9) traffic jam.} \label{figure:different scene}
\end{figure*}

\begin{figure*}[htb]
  \centering{\includegraphics[width=170mm]{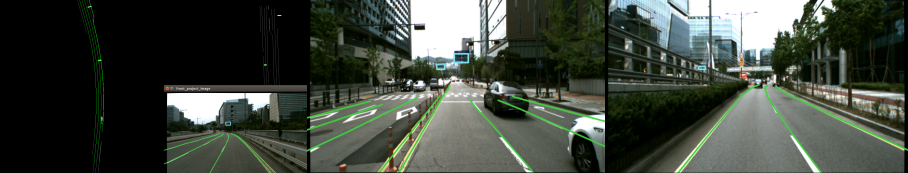}}\\
  \caption{Qualitative results on Kaist dataset. Left: vector format landmark hdmap and projection result. Middle and right are projection results of two different scenes.} \label{figure:kaist}
\end{figure*}

\section{Experiment}
The proposed algorithm is evaluated on two datasets. The first dataset contains around 30 kilometers elevated structured scene in Shanghai, provided by a third party map supplier. The map elements include lane markings, signboard and poles. Due to the compact environment representation in vector format, the storage size of the map is of KB level. The localization system is evaluated in many scenarios, including various weather condition, light intensity and different routes. The second dataset is the public Kaist dataset \cite{jeong2019complex}. Because Kaist dataset does not provide semantic map which is necessary for proposed algorithm, the stereo camera data and high-precision localization pose from Lidar and INS are used to build a semantic landmark map. Qualitative and quantitative experimental results are used to evaluate the accuracy and robustness of the method.

\begin{table*}
\caption{Performance Evaluation on Shanghai dataset}
\centering
\begin{tabular}{c|c|c|ccc|cc|c}
\hline
\multicolumn{1}{c|}{\multirow{2}{*}{Scenario}} & \multicolumn{1}{c|}{\multirow{2}{*}{frame number}} & \multicolumn{1}{c|}{\multirow{2}{*}{Init Success Rate}} & \multicolumn{3}{c|}{3D translation(m)}    & \multicolumn{2}{c|}{2D translation(m)}      & \multicolumn{1}{c}{rotation(deg)} \\ \cline{4-9} 
\multicolumn{1}{c|}{} &\multicolumn{1}{c|}{} &\multicolumn{1}{c|}{} & max  & mean  & \multicolumn{1}{c|}{median} & lateral & \multicolumn{1}{c|}{longitudinal} & \multicolumn{1}{c}{mean}          \\

\hline
sequence1   & 2470 &  $92.20\%$ &1.37       & 0.35       & 0.30         & 0.21        & 0.12       & 0.50           \\
sequence2   & 4312 & $94.24\%$ &  1.22       & 0.46       & 0.43        & 0.24        & 0.23   & 0.43          \\
sequence3   & 7490  & $85.46\%$ & 1.42       & \textbf{0.15}       & \textbf{0.13}        & \textbf{0.09}    & \textbf{0.07}     & \textbf{0.38}          \\
sequence4   & 3596 & $88.79\%$ & 1.22       & 0.35       & 0.32        & 0.21       & 0.17    & 0.47          \\
sequence5   & 2289 & $73.64\%$  & 0.63       & 0.29       & 0.26        & 0.22       & 0.05    & 0.51          \\
sequence6   & 11265 & $89.40\%$ & 2.59       & 0.35       & 0.30        & 0.21      & 0.18     & 0.46          \\
sequence7   & 9297 & $90.93\%$ & 1.34       & 0.33       & 0.29        & 0.20      & 0.16     & 0.64         \\
\hline
    \end{tabular}
    \begin{tablenotes} 
		\item $\star$ This initialization success rate equals initialization success frame within 10 frames / total frames in a sequence. \\
		\item $\star$ The camera of Sequence 3 is a wide angle camera with 120 degree FOV. Others are with 42.5 FOV. \\
     \end{tablenotes}
    
    \label{tab:accuracy}
\end{table*}

\subsection{Qualitative Result}

A accurate localization ensures that the projection of map elements on image is completely consistent with the semantic perception (e.g, Fig. \ref{figure:pose vibration} (a)).  Figure. \ref{figure:pose vibration} lists examples of alignment between HD map and image semantic segmentation. Results of (b)-(g) are the alignments under small perturbations from vehicle pose used in (a). The amount of angle perturbation is 2 degrees and translation perturbation is 1 meter.



We can see that the projection of HD landmarks changes largely with perturbations of pitch, yaw, y and z. In contrast, the projection results are less influenced by  roll angle and vehicle forward direction perturbations.
Also, since the imaging scale is closely related to the vehicle height, lane markings projection will expand to image boundary or shrink to the image center with the wrong vehicle height. Therefore, roll angle and vehicle longitudinal position are not included in optimization stage if there are no signboards or poles present.

Projection results of Shanghai dataset and Kaist dataset are shown in Figure. \ref{figure:different scene} and \ref{figure:kaist}.
For instance, results of different scenes are included in Fig. \ref{figure:different scene}: (1-2) curve road; (3) long straight road in sunny day; (4) rainy; (5) windshield wiper blocks part of the image; (6) diverging ramp; (7) low illumination; (8-9) traffic jam. Our vision localization system achieves robust results on these scenarios.


\subsection{Quantitative Evaluation}
Because of the encryption issue, the third-party map can not exactly match the high-precision trajectory of Novatel, a GNSS inertial navigation system. The relative pose error (RPE) is used as the evaluation metric of localization accuracy. Due to that the lateral and longitudinal localization accuracy are more critical than other metrics. These two errors are reported in the experiments. We use EVO \cite{grupp2017evo} as the accuracy evaluation tool. The frame interval is set to 5, about 12 meter with image frequency of 10Hz and the vehicle speed is around 80 km/h. 

\begin{figure}[]
  \centering{\includegraphics
  [width=60mm]{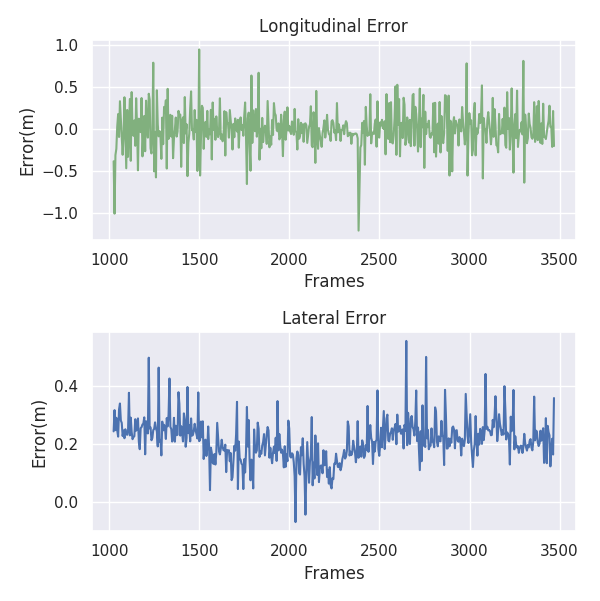}}\\
  \caption{Lateral and longitudinal localization error of Sequence 1.} \label{figure:latlon error}
\end{figure}

Localization accuracy evaluations of several data sequences are reported in Table. \ref{tab:accuracy} and Figure. \ref{figure:latlon error}. In the experiments, wide camera (FOV of 120 degree) is used in sequence 3 and other sequences are with camera having FOV of 42.5 degree. Mean rotation error is below 1 degree. Lateral error and longitudinal error are about 20 cm.  
If the localization is successful within 10 frames by initializing from any frame in sequence, this frame will be labeled as initialization success frame. The initialization success rate can be calculated through manual labeling and observation. Our proposed localization initialization strategy achieves about 90 percent success rate. The failure of initialization is mainly due to the poor GPS signal caused by signal blocking.
In addition, localization accuracy of sequence 3 is improved because more semantic elements are captured in lateral direction by wide angle camera.  

\subsection{Running Time Analysis}

Table. \ref{tab:running time} shows the running time of different modules (without segmentation) in the tracking stage. KD-Tree is used for querying landmarks from urban scale map to improve searching efficiency. The most time-consuming step is the image post-processing, i.e. the process of gradient field construction. With machine setting of i7-8700k CPU and GTX 1050 TI GPU, localization running frequency is near 100Hz. 

\begin{figure}[t]
  \centering{\includegraphics[width=70mm]{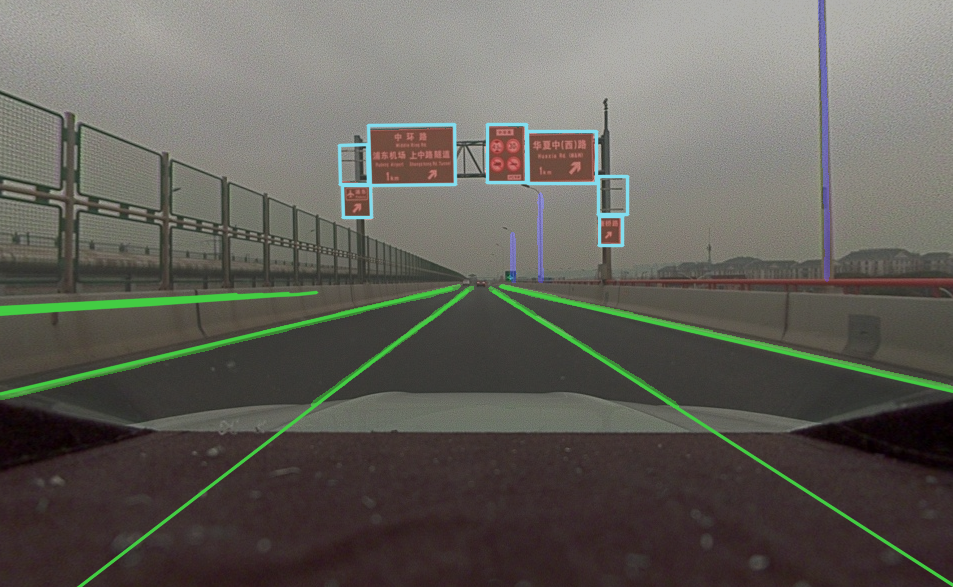}}\\
  \caption{Example of that the HD map is not updated timely. For example, several traffic signs (top-left and top-right rectangles in cyan) are missing in the scene.} \label{figure:map change}
\end{figure}

\begin{figure}[t]
  \centering{\includegraphics[width=85mm]{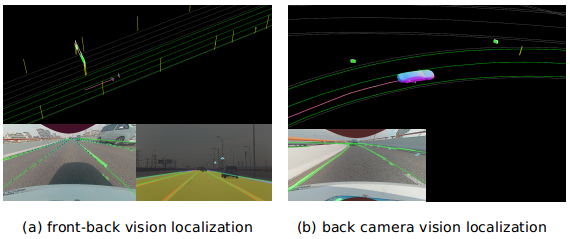}}\\
  \caption{Examples of multi-camera (e.g., front and rear) vision localization. (a) Both front and rear cameras are present. (b ) Front camera is disable in the system.} \label{figure:front back localization}
\end{figure}

\begin{table}\small
\caption{Running time statistics.}
\centering
\begin{tabular}{l|c}
\hline
module & time(ms) \\
\hline
    image post-process &6.24 \\
    map query & 0.73 \\
    pose optimization & 1.76 \\
    pose search &0.60 \\
    total track time & 12.05 \\
\hline
    \end{tabular}
\label{tab:running time}
\end{table}

\subsection{Change of the Scenario}
The HD map usually can not be update timely. While our proposed algorithm is robust to small-scale urban environment change. Also, our method is able to determine the change region of the map, which is significant to the localization and mapping applications. Figure. \ref{figure:map change} shows the signboard layout is changed in the driving scene. Our approach can (1) achieve robust localization and (2) report updated region in the map according to the misalignment between camera and the map.


\subsection{Multi-cameras Support}

The front camera with field of view of 42.5 degree and rear fisheye camera (FOV of 195 degree) are used as our sensor setup of multi-camera localization experiment. 
In order to simplify the calculation, a raw fisheye image is transformed into a pinhole image. Figure. \ref{figure:front back localization} illustrates the localization results by using both cameras and only using the rear camera to simulate that the front camera is disable. It shows that even if the forward looking camera is out of function, successful localization result can still be obtained. As a result, the multi-cameras setup improves the robustness and the accuracy of a localization system.


\section{CONCLUSIONS}

In this paper, we propose a vision-based localization system by using vehicle wheel odometry, ordinary car-equipped consumer level GPS, HD Map and cameras. According to the formulation, our system is able to handle both monocular and multi-camera sensor settings. We also demonstrate that our system is robust to different environment conditions and the change of driving scenarios, and achieves accurate localization results. In future work, we will introduce IMU into localization system to build a visual inertial odometry and GNSS-IMU inertial navigation system. The pose output of VIO system and INS system will be effectively fused with the localization result of our algorithm to form a practical low-cost mass-production localization system. 

\section*{Acknowledgement}
Pengpeng Liang is supported by National Natural Science Foundation of China  (Grant No. 61806181).

\addtolength{\textheight}{-12cm}   




\bibliographystyle{IEEEtran}
\bibliography{refs}

\end{document}